\documentclass[10pt,twocolumn,letterpaper]{article}

\usepackage{cvpr}
\usepackage{times}
\usepackage{epsfig}
\usepackage{graphicx}
\usepackage{amsmath}
\usepackage{amssymb}

\usepackage{array}
\usepackage{cite}

\usepackage{algorithm}
\usepackage{algorithmic}
\usepackage{mdwmath}

\usepackage{graphicx}         
\usepackage{subfigure}
\usepackage{pifont}
\newcommand{\cmark}{\ding{51}} 

\usepackage{bbm}
\usepackage{dsfont}

\usepackage{bm}
\usepackage{comment}
\usepackage{amsmath}
\usepackage{enumerate}  
\usepackage{mathrsfs}
\usepackage{makeidx}         
\usepackage{graphicx}        
\usepackage{subfigure}
\usepackage{epsfig,amssymb,latexsym}
\usepackage{psfrag}
\usepackage{algorithmic}
\usepackage{algorithm}

\usepackage[misc]{ifsym}

\usepackage{fancyhdr}
\usepackage{multirow}

\usepackage{color}
\usepackage{indentfirst}

\newcommand{\mr}{\mathrm}
\newcommand{\mc}{\mathcal}

\usepackage[pagebackref=true,breaklinks=true,letterpaper=true,colorlinks,bookmarks=false]{hyperref}

\cvprfinalcopy 

\def\cvprPaperID{966} 

\ifcvprfinal\fi

\begin{document}

\title{What Can Be Transferred: Unsupervised Domain Adaptation for \\Endoscopic Lesions Segmentation}

\author{
	Jiahua~Dong\textsuperscript{1,2,3},
	Yang Cong\textsuperscript{1,2}\thanks{The corresponding authors are Prof. Yang Cong and Dr. Gan Sun.}~,
	Gan Sun\textsuperscript{1,2$*$},
	Bineng Zhong\textsuperscript{4}
	and Xiaowei Xu\textsuperscript{5} \\	
	\textsuperscript{1}State Key Laboratory of Robotics, Shenyang Institute of Automation, \\ Chinese Academy of Sciences, Shenyang, 110016, China.\thanks{This work is supported by Ministry of Science and Technology of the People´s Republic of China (2019YFB1310300) and NSFC under grant (61722311, U1613214, 61821005, 61533015).}  \\	
	\textsuperscript{2}Institutes for Robotics and Intelligent Manufacturing, \\ Chinese Academy of Sciences, Shenyang, 110016, China. \\
	\textsuperscript{3}University of Chinese Academy of Sciences, Beijing, 100049, China. \\
	\textsuperscript{4}Huaqiao University, Xiamen, Fujian, 361021, China. \\
	\textsuperscript{5}Department of Information Science, University of Arkansas at Little Rock, Arkansas, USA. \\
	{\tt\small dongjiahua@sia.cn, \{congyang81,~sungan1412\}@gmail.com, bnzhong@hqu.edu.cn, xwxu@ualr.edu}
}


\maketitle
\thispagestyle{empty}

\begin{abstract}
Unsupervised domain adaptation has attracted growing research attention on semantic segmentation. However, 1) most existing models cannot be directly applied into lesions transfer of medical images, due to the diverse appearances of same lesion among different datasets; 2) equal attention has been paid into all semantic representations instead of neglecting irrelevant knowledge, which leads to negative transfer of untransferable knowledge. To address these challenges, we develop a new unsupervised semantic transfer model including two complementary modules (\emph{i.e.,} $\mathcal{T}_D$ and $\mathcal{T}_F$) for endoscopic lesions segmentation, which can alternatively determine where and how to explore transferable domain-invariant knowledge between labeled source lesions dataset (\emph{e.g.,} gastroscope) and unlabeled target diseases dataset (\emph{e.g.,} enteroscopy). Specifically, $\mathcal{T}_D$ focuses on where to translate transferable visual information of medical lesions via residual transferability-aware bottleneck, while neglecting untransferable visual characterizations. Furthermore, $\mathcal{T}_F$ highlights how to augment transferable semantic features of various lesions and automatically ignore untransferable representations, which explores domain-invariant knowledge and in return improves the performance of $\mathcal{T}_D$. To the end, theoretical analysis and extensive experiments on medical endoscopic dataset and several non-medical public datasets well demonstrate the superiority of our proposed model. 

\end{abstract}

\section{Introduction}
The successes of unsupervised domain adaptation have been widely-extended into a large amount of computer vision applications, \emph{e.g.,} semantic segmentation \cite{Tsai_2019_ICCV, Dong_2019_ICCV}. Due to the powerful generalization capacity for segmentation task of unlabeled target data, enormous unsupervised domain adaptation methods \cite{Lee_2019_CVPR, Li_2019_CVPR, Lian_2019_ICCV, exp:LtA, exp:CGAN} has been developed to narrow the distribution divergence between labeled source dataset and unlabeled target dataset.

\begin{figure}[t]
	\small
	\centering
	\includegraphics[trim = 9mm 47mm 9mm 57mm, clip, width=240pt, height  =95pt]{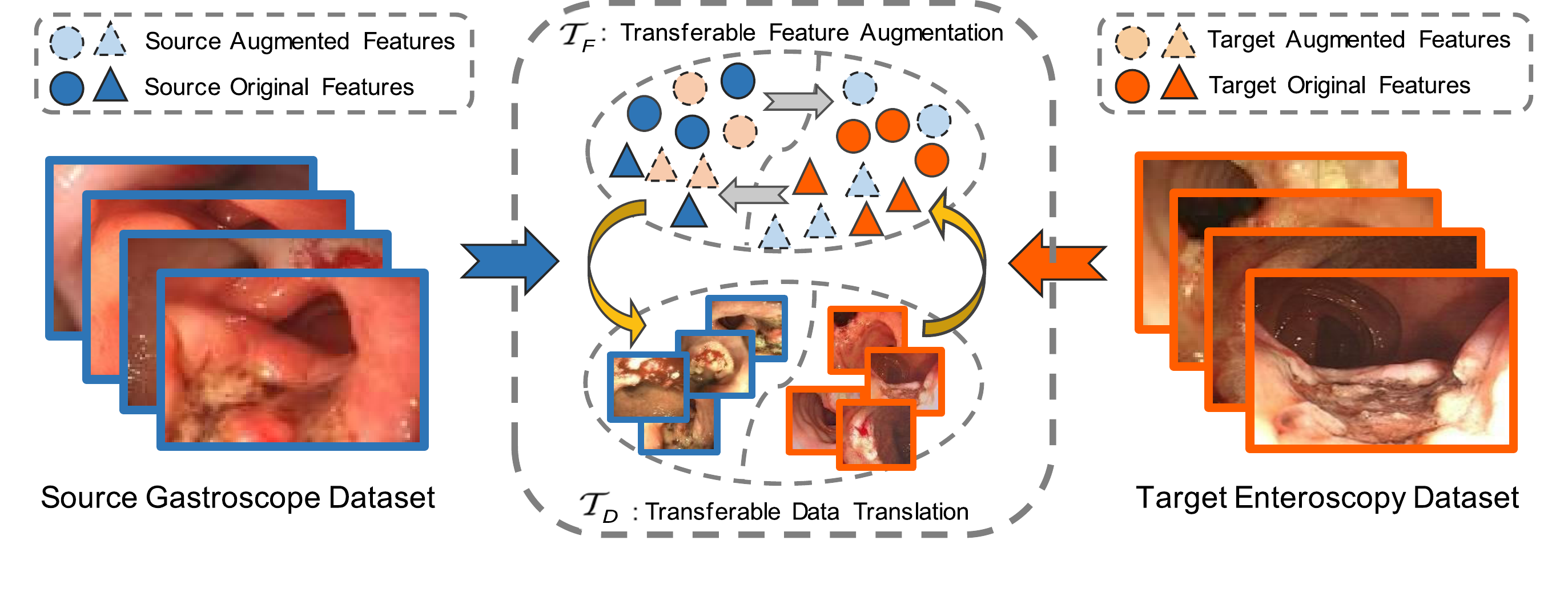}
	\vspace{-20pt}
	\caption{Illustration of our unsupervised semantic transfer model, where two complementary modules $\mc{T}_F$ and $\mc{T}_D$ can alternatively explore where to translate transferable visual characterizations of medical lesions and how to augment transferable semantic feature of various diseases, respectively.}
	\label{fig:demonstration}
\end{figure}

However, most state-of-the-art models \cite{Luo_2019_CVPR, Dong_2019_ICCV, Dou2018UCD3304415, Chen2019SynergisticIA} cannot efficiently address semantic transfer of medical lesions with various appearances, due to the difficulty in determining what kind of visual characterizations could boost or cripple the performance of semantic transfer. Additionally, they fail to brush untransferable representations aside while forcefully utilizing these irrelevant knowledge heavily degrades the transfer performance. \emph{Take the clinical lesions diagnosis as an example, cancer and ulcer present diverse visual information (\emph{e.g.}, appearance, shape and texture) among gastroscope and enteroscopy datasets, which is a thorny diagnosis challenge due to the large distribution shift among different datasets. Obviously, it is difficult to manually determine what kind of lesions information could promote the transfer performance, \emph{i.e.}, exploring domain-invariant knowledge for various lesions.} Therefore, how to automatically capture transferable visual characterizations and semantic representations while neglecting irrelevant knowledge across domains is our focus in this paper.

To address the above mentioned challenges, as shown in Figure~\ref{fig:demonstration}, we develop a new unsupervised semantic lesions transfer model to mitigate the domain gap between labeled source lesions dataset (\emph{e.g.}, gastroscope) and unlabeled target diseases dataset (\emph{e.g.}, enteroscopy). To be specific, the proposed model consists of two complementary modules, \emph{i.e.}, $\mathcal{T}_D$ and $\mathcal{T}_F$, which could automatically determine where and how to explore transferable knowledge from source diseases dataset to assist target lesions segmentation task. On one hand, motivated by information theory \cite{45903}, residual transferability-aware bottleneck is developed for $\mathcal{T}_D$ to highlight where to translate transferable visual information while preventing irrelevant translation. On the other hand, Residual Attention on Attention Block ($\mr{RA^2B}$) is proposed to encode domain-invariant knowledge with high transferability scores, which assists $\mathcal{T}_F$ in exploring how to augment transferable semantic features and boost the translation performance of module $\mathcal{T}_D$ in return. Meanwhile, target samples are progressively assigned with confident pseudo pixel labels along the alternative training process of $\mathcal{T}_D$ and $\mathcal{T}_F$, which further bridges the distribution shift in the retraining phase. Finally, theoretical analysis about our proposed model in term of narrowing domain discrepancy among source and target datasets is elaborated. Extensive experiments on both medical endoscopic dataset and several non-medical datasets are conducted to justify the effectiveness of our proposed model.

The main contributions of this paper are as follows:
\begin{itemize}
	\vspace{-8pt}
	\setlength{\itemsep}{0pt}		
	\setlength{\parsep}{0pt}	
	\setlength{\parskip}{0pt}
\item A new unsupervised semantic representations transfer model is proposed for endoscopic lesions segmentation. To our best knowledge, this is an earlier attempt to automatically highlight the transferable semantic knowledge for endoscopic lesions segmentation in the biomedical imaging field.
	
	\item Two complementary modules $\mathcal{T}_D$ and $\mathcal{T}_F$ are developed to alternatively explore the transferable representations while neglecting untransferable knowledge, which can not only determine where to translate transferable visual information via $\mathcal{T}_D$, but also highlight how to augment transferable representations via $\mathcal{T}_F$.
	
	\item Comprehensive theory analysis about how our model narrows domain discrepancy is provided. Experiments are also conducted to validate the superiority of our model against state-of-the-arts on the medical endoscopic dataset and several non-medical public datasets.
	
\end{itemize}

\section{Related Work} \label{sec:related work}
This section reviews some related works about semantic lesions segmentation and unsupervised domain adaptation.

\textbf{Semantic Segmentation of Lesions:}
Deep neural networks \cite{wang2019laplacian, wang_TMM} have achieved significant successes in enormous applications, \emph{e.g.}, medical lesions segmentation \cite{Bozorgtabar2017, dezsampx001512018, xuLargeScaleTissue2017, article_Baillard, Dong_2019_ICCV}. When compared with traditional models \cite{ChengComputerAided, HorschAutomaticSeg} requiring handcrafted lesions features, it relies on powerful lesions characterization capacity to boost accuracy and efficiency of diseases diagnosis, but needs large-scale pixel labels. To save the annotations cost, unsupervised learning has been widely-applied into medical lesions segmentation \cite{DBLP-journals/corr/abs-1806-04972, Dou2018UCD3304415, Bozorgtabar2017, BowlesBrainLesion, Atlason2018UnsupervisedBL, pmlr-v102-baur19a}. However, these models require effective prior information \cite{DBLP-journals/corr/abs-1806-04972} or distribution hypothesis \cite{Atlason2018UnsupervisedBL} to generalize previously unseen diseases, which only produces inaccurate and coarse lesions prediction. Thus, it is a thorny challenge to perform well on unseen target lesions when training on source diseases data \cite{Dou2018UCD3304415, Chen2019SynergisticIA, Dong_2019_ICCV}.

\textbf{Unsupervised Domain Adaptation:}
After Hoffman \emph{et al.}~\cite{exp:Wild} first utilize adversarial network \cite{Goodfellow:2014:GAN} to achieve domain adaptation for semantic segmentation task, diverse variants based on adversarial strategy \cite{exp:LtA, exp:CCA, exp:LSD, exp:CGAN, Wu_2018_ECCV, Saito_2018_CVPR} are proposed to address the domain shift challenge. Different from these models, \cite{exp:CL, Lian_2019_ICCV} employ curriculum learning to infer important properties for target images according to source samples. \cite{Zou_2018_ECCV} designs a non-adversarial model to transfer semantic representation in a self-training manner. \cite{Gong_2019_CVPR} presents the domain flow translation to explore expected intermediate domain. Li \emph{et al.} \cite{Li_2019_CVPR} propose a bidirectional learning model for target adaptation. In addition, novel adaptation losses \cite{Lee_2019_CVPR, Vu_2019_CVPR, Luo_2019_CVPR} are designed to measure discrepancy among different datasets. \cite{Dong_2019_ICCV} develop a pseudo pixel label generator to focus on hard-to-transfer target samples. \cite{Tsai_2019_ICCV, Luo_2019_ICCV, ding2018robust, ding2018graph, NIPS2019_8940} explore discriminative semantic knowledge to narrow the distribution divergence.

%
%

\begin{figure*}[t]
\small
\centering
\includegraphics[trim = 16mm 65mm 0mm 70mm, clip, width =503pt, height =138pt]{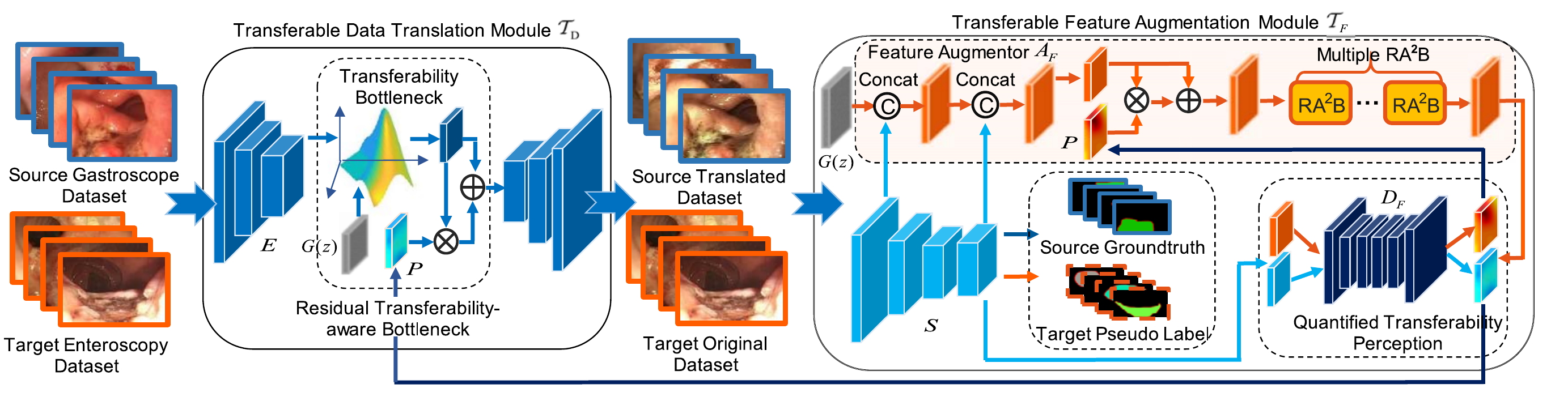}
\vspace{-20pt}
\caption{Overview architecture of our proposed model, which is composed of two alternatively complementary modules $\mathcal{T}_D$ and $\mathcal{T}_F$. Specifically, $\mathcal{T}_D$ focuses on exploring where to translate transferable visual characterizations via residual transferability-aware bottleneck. $\mathcal{T}_F$ highlights how to augment transferable semantic representations while neglecting untransferable knowledge, which incorporates multiple residual attention on attention blocks ($\mr{RA^2B}$) to capture domain-invariant features with high transferability.}
\label{fig:overview_framework} 
\vspace{-10pt}
\end{figure*}

\section{The Proposed Model}
In this section, we first present overall framework of our proposed model and then introduce detailed model formulation, followed by comprehensive theoretical analysis. 

\subsection{Overview}
Given the source dataset (\emph{e.g.,} gastroscope) $X_s = \{x_i^s, y_i^s\}_{i=1}^m$  and target dataset (\emph{e.g.,} enteroscopy) $X_t = \{x_j^t\}_{j=1}^n$, where $x_i^s$ and $x_j^t$ represent source samples with pixel annotations $y_i^s$ and target images without pixel labels, respectively. Although existing semantic transfer models \cite{Lee_2019_CVPR, Vu_2019_CVPR, Luo_2019_CVPR, Tsai_2019_ICCV, Luo_2019_ICCV} attempt to narrow the distribution shift among source and target datasets, semantic representations are not all transferable while forcefully taking advantage of irrelevant knowledge could lead to negative transfer. Besides, various lesions with diverse appearances make them difficult to explore what kind of visual characterizations will promote transfer performance. Therefore, we endeavor to automatically highlight the transferable representations between source and target datasets to improve the lesions segmentation performance for unlabeled target samples, while ignoring the irrelevant knowledge for semantic transfer.

As depicted in Figure~\ref{fig:overview_framework}, the proposed model consists of two complementary modules, \emph{i.e.}, $\mathcal{T}_D$ and $\mathcal{T}_F$, which alternatively determines where and how to highlight transferable knowledge. Specifically, with quantified transferability perception from discriminator $D_F$ in $\mathcal{T}_F$, the source samples $x_i^s$ are first passed into $\mathcal{T}_D$ to explore where to translate transferable visual characterizations, according to the style information of target images $x_j^t$. Afterwards, we forward the translated source samples $\hat{x}_i^s$ along with $x_j^t$ into $\mathcal{T}_F$ to determine how to augment transferable semantic features while ignoring those untransferable representations. $\mathcal{T}_F$ further mitigates the domain gap in the feature space and in return promotes the translation performance of $\mathcal{T}_D$. Our model could be regarded as a closed loop to alternatively update the parameters of $\mathcal{T}_D$ and $\mathcal{T}_F$. Furthermore, along the alternative training process of $\mathcal{T}_D$ and $\mathcal{T}_F$, our model progressively mines confident pseudo pixel labels $\hat{y}_j^t$ for target samples, which fine-tunes the segmentation model $S$ in $\mathcal{T}_F$ to learn domain-invariant knowledge.

\subsection{Quantified Transferability Perception} \label{sec:quantified transferability}
Intuitively, domain uncertainty estimation of the discriminator $D_F$ in $\mathcal{T}_F$ can assist in identifying those representations which can be transferred, cannot be transferred, or already transferred. For example, the input source features $F_s$ and target features $F_t$ that are already aligned across domains will fool the discriminator $D_F$ for distinguishing whether the input is from $X_s$ or $X_t$. In other words, we can easily discriminate whether the input feature maps $F_s$ or $F_t$ is transferable or not according to the output probabilities of discriminator $D_F$. Therefore, in order to highlight those transferable representations, we utilize uncertainty measure function of information theory (\emph{i.e.,} entropy criterion $\mc{I}(p) = -\sum_r p_r \mathrm{log}(p_r)$) to quantify the transferability perception of corresponding semantic features. Take the source samples as an example, given the output probability $D_F(F_s; \theta_{D_F})$ of discriminator $D_F$ with network weights $\theta_{D_F}$, the transferability perception for input source feature $F_s$ can be formally quantified as follows:
\begin{equation}
P = 1 - \mc{I}(D_F(F_s; \theta_{D_F})).
\label{equ:transferability}
\end{equation}
Similarly, Eq.~\eqref{equ:transferability} can also quantify the transferability for target features $F_t$ according to the output $D_F(F_t; \theta_{D_F})$. Note that the quantified transferability for source and target features share the same notation $P$ for simplification.

However, false transferability perception may hurt semantic transfer task to some degree. Therefore, residual transferability perception mechanism is designed to feedback the positive transferability into $\mathcal{T}_D$ in Section \ref{sec:data_translation} and feature augmentor $A_F$ in Section \ref{sec:feature_transfer}, as shown in Figure~\ref{fig:overview_framework}.

\subsection{Transferable Data Translation ($\mathcal{T}_D$)} \label{sec:data_translation}
Different from previous translation model \cite{domain:class-preserve}, our module $\mathcal{T}_D$ could highlight where to translate transferable visual characterizations for better transfer performance. With the quantified transferability perception from $D_F$, $\mathcal{T}_D$ pays more attention to selectively explore transferable mappings $X_s \rightarrow X_t$ and $X_t \rightarrow X_s$ while preventing irrelevant translations with low transfer scores. The samples from both $X_s$ and $X_t$ are forwarded into $\mathcal{T}_D$ to train the translation model, which produces the corresponding translated source dataset $\hat{X}_s=\{\hat{x}_i^s, y_i^s\}_{i=1}^m$ and mapped target dataset $\hat{X}_t=\{\hat{x}_j^t\}_{j=1}^n$. $\hat{x}_i^s = \mathcal{T}_D(x_i^s; \theta_{\mathcal{T}_D})$ and $\hat{x}_j^t = \mathcal{T}_D^{-1}(x_j^t; \theta_{\mathcal{T}_D^{-1}})$ correspond to translated samples from $\hat{X}_s$ and $\hat{X}_t$, where $\theta_{\mathcal{T}_D}$ and $\theta_{\mathcal{T}_D^{-1}}$ are network parameters of $\mathcal{T}_D$ and $\mathcal{T}_D^{-1}$, respectively, and $\mathcal{T}_D^{-1}$ denotes the reverse translation of $\mathcal{T}_D$ that learns the mapping $X_t \rightarrow X_s$. Notice that translated source images $\hat{x}_i^s$ share same pixel annotations $y_i^s$ with original image $x_i^s$, though there exists large visual gap among them. To encourage $\hat{X}_s$ have closer distribution with $X_t$, $\mathcal{L}_{d}^a(\hat{X}_s, X_t)$ is employed to train $\mathcal{T}_D$, which can be written as follows:
\begin{equation}
\begin{split}
\mathcal{L}_{d}^a(\hat{X}_s,& X_t) = \mathbb{E}_{x_{j}^t \in X_t}\big[\mr{\log} (D_1(x_j^t; \theta_{D_1}))\big] + \\
&\mathbb{E}_{x_{i}^s \in X_s} \big[1 - \mr{log}(D_1(\mathcal{T}_D(x_i^s; \theta_{\mathcal{T}_D}); \theta_{D_1})) \big],
\end{split}
\label{equ:loss_adv_target}  	
\end{equation}
where $D_1$ is the discriminator with network parameters $\theta_{D_1}$ that distinguishes between translated source images $\hat{x}_i^s$ and real target samples $x_j^t$. Likewise, we utilize $\mathcal{L}_{d}^a(X_s, \hat{X_t})$ to learn the mapping translation from $X_t$ to $X_s$, \emph{i.e.}, 
\begin{equation}
\begin{split}
\mathcal{L}_{d}^a(& X_s, \hat{X}_t) = \mathbb{E}_{x_{i}^s \in X_s}\big[\mr{\log} (D_2(x_i^s; \theta_{D_2}))\big] + \\
&\mathbb{E}_{x_{j}^t \in X_t} \big[1 - \mr{log}(D_2(\mathcal{T}_D^{-1}(x_j^t; \theta_{\mathcal{T}_D^{-1}}); \theta_{D_2})) \big],
\end{split}
\label{equ:loss_adv_source}  	
\end{equation}
where $D_2$ shares similar definition with $D_1$ but discriminates whether the inputs are from real source images $x_i^s$ or translated target samples $\hat{x}_j^t$. $\theta_{D_2}$ represents the corresponding network weights of $D_2$. Additionally, semantic consistency between input and reconstructed samples for both source and target data are ensured by the loss $\mathcal{L}_{d}^c(X_s, X_t)$:
\begin{equation}
\begin{split}
\mathcal{L}_{d}^c(X_s, X_t) & = \mathbb{E}_{x_{j}^t \in X_t} \big[ \left\|\mathcal{T}_D(\hat{x}_j^t; \theta_{\mathcal{T}_D}) - x_j^t \right\|_1 \big] +  \\
&\mathbb{E}_{x_{i}^s \in X_s}\Big[ \left\| \mathcal{T}_D^{-1}(\hat{x}_i^s; \theta_{\mathcal{T}_D^{-1}}) - x_i^s \right\|_1 \Big].
\end{split}
\label{equ:loss_cycle_consistent}  	
\end{equation}
As a result, the overall objective $\mathcal{L}_{\mathcal{T}_D}$ for training $\mathcal{T}_D$ is:
\begin{equation}
\begin{split}
\mathcal{L}_{\mathcal{T}_D} = \mathcal{L}_{d}^a(\hat{X}_s, X_t)+\mathcal{L}_{d}^a(X_s, \hat{X}_t) + \alpha \mathcal{L}_d^c(X_s, X_t).
\end{split}
\label{equ:objective_Mv} 	
\end{equation}

However, Eq.~\eqref{equ:objective_Mv} cannot selectively capture important semantic knowledge with high transferability. Therefore, as shown in Figure~\ref{fig:overview_framework}, we develop a residual transferability-aware bottleneck, which determines where to translate transferable information by purifying semantic knowledge with high transfer scores. Specifically, built upon the information theory \cite{45903}, we design an information constraint on the latent feature space, which is adaptively weighted by the quantified transferability perception $P$ in Eq.~\eqref{equ:transferability}. It encourages the feature extractor $E$ in $\mathcal{T}_D$ to encode transferable representations. Formally, Eq.~\eqref{equ:objective_Mv} can be reformulated as:
\begin{equation}
\begin{split}
&\mathcal{L}_{\mathcal{T}_D} = \mathcal{L}_d^a(\hat{X}_s, X_t)+\mathcal{L}_d^a(X_s, \hat{X}_t) + \alpha \mathcal{L}_d^c(X_s, X_t),   \\
&\quad s.t.~ \mathbb{E}_{x_{i}^s \in X_s} \big[P\odot\mr{KL}(E(x_i^s; \theta_E)||G(z))\big] \leq T_s , \\
&\quad\quad ~~ \mathbb{E}_{x_{j}^t \in X_t} \big[P\odot\mr{KL}(E(x_j^t; \theta_E)||G(z))\big] \leq T_t ,
\end{split}
\label{equ:overall_objective_Mv} 	
\end{equation}
where $\odot$ represents the channel-wise product. $G(z)$ is the marginal distribution of $z$, which denotes the standard Gaussian distribution $\mathcal{N}(0; I)$. $T_s$ and $T_t$ represent transferability bottleneck thresholds for source and target datasets, respectively. They are set as the same value in this paper and denoted as $T$ for simplification. $E(x_i^s; \theta_E)$ and $E(x_j^t; \theta_E)$ are the extracted features via $E$ for source and target samples, where $\theta_E$ denotes the network parameters. Take samples $x_i^s$ as the intuitive explanation for Eq.~\eqref{equ:overall_objective_Mv}: the larger KL divergence among $E(x_i^s; \theta_E)$ and $G(z)$ indicates the closer dependence among $x_i^s$ and $z$, which enforces $z$ to encode more semantic representations from samples $x_i^s$. Obviously, these semantic representations are not all transferable for translation while utilizing irrelevant knowledge leads to the negative transfer. Thus, by enforcing KL divergence weighted with quantified transferability $P$ to the threshold $T$, untransferable representations from $G(z)$ could be neglected, which is then regarded as latent feature of $x_i^s$ and forwarded into the decoder network. To optimize Eq.~\eqref{equ:overall_objective_Mv}, we equally formulate it as Eq.~\eqref{equ:Mv_Lagrange} by employing two Lagrange multipliers $\lambda_s$ and $\lambda_t$ for source and target datasets:
\begin{equation}
\begin{split}
&\mathcal{L}_{\mathcal{T}_D} =  \mathcal{L}_d^a(\hat{X}_s, X_t)+\mathcal{L}_d^a(X_s, \hat{X}_t) +  \alpha \mathcal{L}_d^c(X_s, X_t) \\
&~~ + \lambda_s \big( \mathbb{E}_{x_{i}^s \in X_s}  \big[P\odot(\mr{KL}(E(x_i^s; \theta_E)||G(z)) - T)\big] \big)   \\
&~~ + \lambda_t \big( \mathbb{E}_{x_{j}^t \in X_t} \big[P\odot(\mr{KL}(E(x_j^t; \theta_E)||G(z)) - T) \big] \big), 
\end{split}
\label{equ:Mv_Lagrange}
\end{equation}
where $\lambda_s$ and $\lambda_t$ are updated by $\lambda_s \leftarrow \mr{max}(\lambda_s, \gamma \mathcal{L}_b^s)$ and $\lambda_t \leftarrow \mr{max}(\lambda_t, \gamma \mathcal{L}_b^t)$, respectively. The last two terms of Eq.~\eqref{equ:Mv_Lagrange} are defined as the transferability constraint losses $\mathcal{L}_b^s$ and $\mathcal{L}_b^t$. $\gamma$ denotes the updating step of $\lambda_s$ and $\lambda_t$. 

\subsection{Transferable Feature Augmentation ($\mathcal{T}_F$)} \label{sec:feature_transfer}
Although $\mathcal{T}_D$ is designed to translate transferable visual characterizations, it cannot ensure feature distribution across domains to be well aligned. Motivated by this observation, transferable feature augmentation module $\mathcal{T}_F$ is developed to automatically determine how to augment transferable semantic features, which further mitigates the domain gap among different datasets and in return boosts the performance of $\mathcal{T}_D$. As depicted in Figure~\ref{fig:overview_framework}, feature augmentor $A_F$ encodes transferable representations from low-level and high-level layers that preserve informative details by incorporating with multiple residual attention on attention blocks ($\mr{RA^2B}$), where $\mr{RA^2B}$ focuses on highlighting the relevance transferability of transferable representations and the details of $\mr{RA^2B}$ are presented as follows.

\begin{figure}[h]
	\vspace{-8pt}
	\small
	\centering
	\includegraphics[trim = 9mm 58mm 9mm 68mm, clip, width=240pt, height  =80pt]{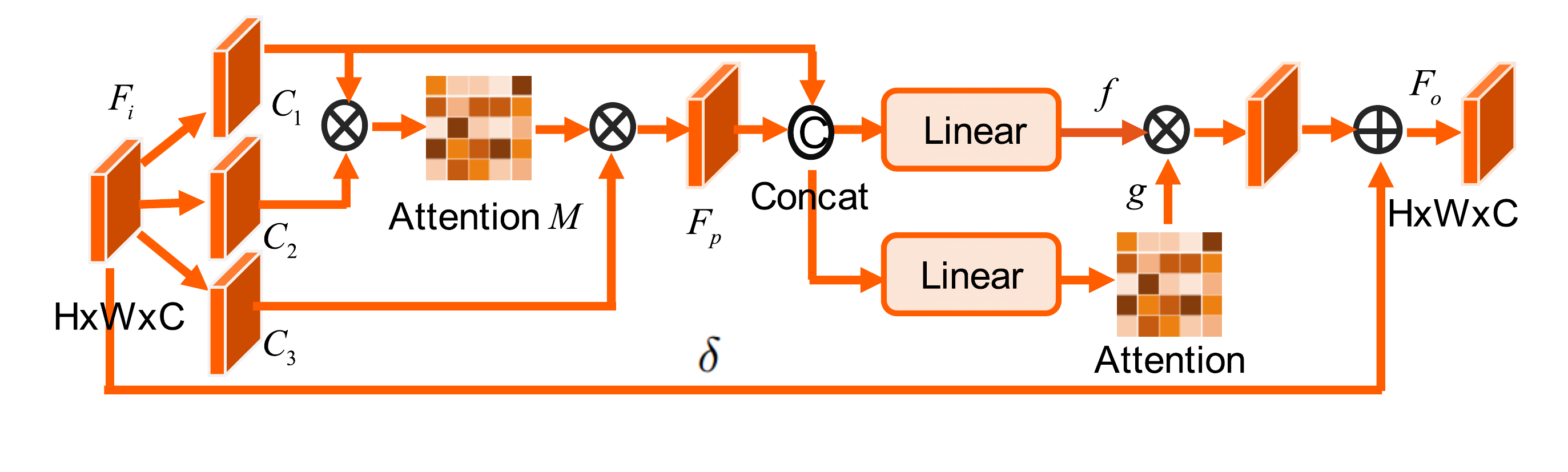}
	\vspace{-24pt}
	\caption{The detailed illustration of $\mr{RA^2B}$.}
	\label{fig:details_RA3B}
	\vspace{-8pt}
\end{figure}

As shown in Figure~\ref{fig:details_RA3B}, given the input feature $F_i\in\mathbb{R}^{H\times W\times C}$, we forward it into three convolutional blocks to produce three new features $C_1, C_2$ and $C_3$ ($C_1, C_2, C_3\in\mathbb{R}^{H\times W\times C}$), where $H, W$ and $C$ represent the height, width and channels of corresponding features. After reshaping $C_1$ and $C_2$ into $\mathbb{R}^{N\times C}$ ($N = H\times W$ denotes the number of pixel positions), attention matrix $M=\{M_{ij}\}_{i,j=1}^N\in\mathbb{R}^{N\times N}$ with softmax activation is obtained. We then utilize the matrix multiplication operator between the transpose of $M$ and reshaped $C_3\in \mathbb{R}^{N\times C}$ to output the attention feature map $F_p\in \mathbb{R}^{N\times C}$, where $M$ and $F_p$ can be formulated as:
\begin{equation}
\hspace{-9pt}M_{ij} = \frac{\mr{exp}\big((C_1C_2^{\top})_{ij}\big)}{\sum_{i=1}^N  \mr{exp}\big((C_1C_2^{\top})_{ij}\big)}, (F_p)_i = \sum_{j=1}^N M_{ij}(C_3)_j, \hspace{-5pt}
\label{equ:position_attention}  	
\end{equation}
where $(F_p)_i$ and $(C_3)_j$ respectively denote the corresponding features at the $i$-th and $j$-th pixel positions. Even though there is no relevant transferable features, Eq.~\eqref{equ:position_attention} still generates an average weighted feature map $F_p$, which could heavily degrade the transferability of semantic knowledge or even encourage them to be untransferable.

Therefore, we develop the $\mr{RA^2B}$ module to measure the relevance between attention result $F_p$ and input feature $C_1$. Then the transferable information flow $f$ and relevance gate $g$ are produced via the linear operation on $F_p$ and $C_1$, \emph{i.e.},
\begin{equation}
\begin{split}
&f = W_f^1C_1 + W_f^2F_p + B_f,   \\
&g = \mr{sigmoid}(W_g^1C_1 + W_g^2F_p + B_g),
\end{split}
\label{equ:atten_on_atten}  	
\end{equation}
where $W_f^1, W_f^2, W_g^1, W_g^2 \in \mathbb{R}^{N \times N}$, $B_f, B_g \in \mathbb{R}^{N \times C}$ are the transformation matrices. Afterwards, $f$ and $g$ are reshaped into $\mathbb{R}^{H\times W\times C}$ and employed to perform element-wise multiplication. We multiply the produced result by a scalar parameter $\delta$ and employ an element-wise sum operation with $F_i$ to obtain the ultimate feature $F_o\in\mathbb{R}^{H\times W\times C}$: 
\begin{equation}
F_o = \delta (f\otimes g) + F_i,
\label{equ:atten_on_atten_feature}  	
\end{equation}
where $\delta$ is initialized as 0, and its value is adaptively learned along the training process.

With quantified transferability perception from $D_F$, $A_F$ could selectively augment the transferable representations while preventing the irrelevant augmentation, which promotes the segmentation module $S$ to learn domain-invariant knowledge and further improves the performance of $\mathcal{T}_D$ in return. The details about how to train $\mathcal{T}_F$ are as follows:

\textbf{Step I:}
The translated source samples $\hat{x}_i^s$ with pixel annotations $y_i^s$ and target images $x_j^t$ with generated pseudo pixel labels $\hat{y}_j^t$ are forwarded into segmentation model $S$, where $\hat{y}_j^t = \mr{argmax}(S(x_j^t; \theta_{S}))$ and $\theta_{S}$ indicates network weights of $S$. The segmentation loss $\mathcal{L}_f^s$ for training $S$ can be concretely expressed as: 
\begin{equation}
\begin{split}
& \min\limits_{\theta_{S}}\mathcal{L}_f^s \!= \! \mathbb{E}_{(\hat{x}_i^s, y_i^s)\in \hat{X}_s}[- \sum\limits_{u = 1}^{\left|\hat{x}_i^s \right|} \sum_{k=1}^{K} \mathbf{1}_{k=(y_i^s)_u}\mr{log}(S(\hat{x}_i^s; \theta_{S})_u^k)]  \\
&\quad~~ + \mathbb{E}_{(x_j^t, \hat{y}_j^t)\in X_t}[- \sum\limits_{v = 1}^{\left|x_j^t \right|} M_v \sum_{k=1}^{K} \mathbf{1}_{k=(\hat{y}_j^t)_v}\mr{log}(S(x_j^t; \theta_{S})_v^k)],
\end{split}
\label{equ:S_training_segmentation}
\end{equation}
where $S(\hat{x}_i^s; \theta_{S})_u^k$ and $S(x_j^t; \theta_{S}))_v^k$ denote the output probabilities of $S$ predicted as class $k$ at the $u$-th and the $v$-th pixels, respectively. $K$ is the classes number. $M_v = \mathbf{1}_{\mr{max}(S(x_j^t; \theta_{S})_v) \ge \beta}$ generates confident pseudo labels at $v$-th pixel for training, where $\beta = 0.9$ is a probability threshold.

\textbf{Step II:}
In order to encourage $A_F$ synthesize new transferable features that resemble the extracted features from source or target datasets (\emph{i.e.}, feature augmentation), $D_F$ is employed to distinguish whether the input is from $A_F$ or $S$. Intuitively, with the assistance of quantified transferability $P$ from $D_F$, $A_F$ selectively augments transferable domain-invariant features while neglecting untransferable knowledge. Consequently, $\mathcal{L}_f^a$ is designed to train $A_F$ while fixing the parameters of $S$ learned from \textbf{Step I}:
\begin{equation}
\begin{split}
&\min\limits_{\theta_{A_F}}\max\limits_{\theta_{D_F}}~~\mathcal{L}_f^a=\mathbb{E}_{x\in (\hat{X}_s, X_t)}[\mr{log}(D_F(S(x; \theta_S); \theta_{D_F}))] + \\
&\quad \mathbb{E}_{x\in (\hat{X}_s, X_t), z\in G(z)}[\mr{log}(1 - D_F(A_F(x, z; \theta_{A_F}); \theta_{D_F}))], 
\end{split}
\label{equ:Af_training}
\end{equation}
where $\theta_{A_F}$ are parameters of $A_F$. $G(z)=\mathcal{N}(0; I)$ denotes Gaussian distribution from which noise samples are drawn. 

\textbf{Step III:} 
The network weights of $A_F$ learned in \textbf{Step II} are fixed in \textbf{Step III}. $D_F$ is retrained to discriminate whether the input is from original datasets or augmented transferable representation. It encourages $S$ to explore a common feature space, where target features are indistinguishable from the source one. As a result, the training objective $\mathcal{L}_f^t$ in Eq.~\eqref{equ:S_training_adversarial} is proposed to optimize $S$, which captures transferable domain invariant knowledge while neglecting the untransferable representations. 
\begin{equation}
\begin{split}
\min\limits_{\theta_S}&\max\limits_{\theta_{D_F}}\mathcal{L}_f^t=\mathbb{E}_{x\in (\hat{X}_s, X_t)}[\mr{log}(1 - D_F(S(x; \theta_S); \theta_{D_F})] \\
& + \mathbb{E}_{x\in (\hat{X}_s, X_t), z\in G(z)}[\mr{log}(D_F(A_F(x, z; \theta_{A_F}); \theta_{D_F}))],
\end{split}
\label{equ:S_training_adversarial}
\end{equation}

Notice that the quantified transferability perception $P$ in Section \ref{sec:quantified transferability} is from $D_F$ in \textbf{Step III} rather than \textbf{Step II}.

\begin{table*}[t]
\centering
\setlength{\tabcolsep}{1.3mm}
\caption{Performance comparison between our proposed model and several competing methods on medical endoscopic dataset.}
\scalebox{0.96}{
\begin{tabular}{|c|ccccccccc|c|}
	\hline
	Metrics & BL \cite{net:deeplab} & LtA \cite{exp:LtA} & CGAN \cite{exp:CGAN} & CLAN \cite{Luo_2019_CVPR} & ADV \cite{Vu_2019_CVPR} & BDL \cite{Li_2019_CVPR} & SWES \cite{Dong_2019_ICCV} & DPR \cite{Tsai_2019_ICCV} & PyCDA \cite{Lian_2019_ICCV} & ~Ours~ \\
	\hline
	\hline
	$\mr{IoU}_n$($\%$) & 74.47  & 81.04 & 79.75 & 81.74 & 81.95 & 84.22 & 83.96 & 83.23 & 84.31 & \textbf{85.48} \\	
	
	$\mr{IoU}_d$($\%$) & 32.65 & 40.35 & 40.52 & 41.33 & 42.27 & 42.84 & 42.63 & 42.11 & 43.08 & \textbf{43.67} \\
	
	mIoU($\%$)         & 53.56 & 60.70 & 60.13 & 61.54 & 62.11 & 63.53 & 63.29 & 62.67 & 63.70 & \textbf{64.58}  \\  		
	\hline					
\end{tabular}
} 				
\label{tab:exp_medical_dataset}
\vspace{-5pt}
\end{table*}

\subsection{Implementation Details}
\textbf{Network Architecture:}
For the transferable visual translation module $\mathcal{T}_D$, CycleGAN \cite{DBLP:journals/corr/ZhuPIE17} is employed as the baseline network. As depicted in Figure~\ref{fig:overview_framework}, the residual transferability-aware bottleneck is attached on the last convolutional block of $\mathcal{T}_D$. In the transferable feature augmentation module $\mathcal{T}_F$, segmentation network $S$ is DeepLab-v3 \cite{net:deeplab} with ResNet-101 \cite{net:resnet} as the backbone architecture, whose the strides of the last two convolutional blocks are transformed from 2 to 1 for higher dimension output. $A_F$ encodes the features from the bottom and the last convolutional blocks of $S$, which are first augmented with the noise from Gaussian distribution.
For discriminator $D_F$, we utilize 5 fully convolutional layers with channel number as \{16, 32, 64, 64, 1\}, where the leaky RELU function parameterized by 0.2 is employed to activate each layer excluding the last convolution filter activated by the sigmoid function.

\textbf{Training and Testing:}
Two complementary modules $\mathcal{T}_D$ and $\mathcal{T}_F$ are alternatively trained until convergence. When training the network $\mathcal{T}_D$, inspired by \cite{DBLP:journals/corr/ZhuPIE17}, we set $\alpha = 10$. The learning rate is initialized as $2.5\times10^{-4}$ for first 10 epochs and linearly decreases to 0 in the later 5 epochs. In Eq.~\eqref{equ:Mv_Lagrange}, $T = 200$, $\lambda_s$ and $\lambda_t$ are initialized as $1.0\times10^{-4}$ with updating step $\gamma$ as $1.0\times10^{-6}$. For backbone DeepLab-v3 \cite{net:deeplab}, we utilize SGD optimizer with an initial learning rate as $2.0\times10^{-4}$ and power as 0.9. The Adam optimizer with initial learning rate as $1.0\times10^{-4}$ is employed for training $D_F$. We set its momentum as 0.9 and 0.99.
In the testing stage, the target images $x_j^t$ (\emph{e.g.}, enteroscopy) are directly forwarded into $S$ for evaluation.

\subsection{Theoretical Analysis} \label{sec:Theoretical_Analysis}
In this subsection, we elaborate the theoretical analysis about our model in term of narrowing domain discrepancy $d_{\mathcal{H}}(P_s, P_t)$ between source and target distributions ($P_s$ and $P_t$), with regard to the hypothesis set $\mathcal{H}$. As pointed out by \cite{BenDavid2010}, the expected error $\epsilon^{P_t}(h)$ of any classifier $h\in\mathcal{H}$ performing on target dataset has theory upper bound, \emph{i.e.},
\begin{equation}
\begin{split}
\forall h\in\mathcal{H}, \epsilon^{P_t}(h) \leq \epsilon^{P_s}(h) + \frac{1}{2} d_{\mathcal{H}}(P_s, P_t) + \Gamma,
\end{split}
\label{equ:error_upper_bound}
\end{equation}
where $\Gamma$ is an independent constant. $\epsilon^{P_s}(h)$ is the expected error of any $h\in\mathcal{H}$ classifying on source samples, which can be negligibly small under the supervisory training. $d_{\mathcal{H}}(P_s, P_t) = 2\sup\limits_{h\in\mathcal{H}}\big|\Pr\limits_{x_i^s\sim P_s}[h(x_i^s)=1] - \Pr\limits_{x_j^t\sim P_t}[h(x_j^t)=1]\big|$ denotes the $\mathcal{H}$-divergence distance between $P_s$ and $P_t$. Thus, the relationships between our model and domain discrepancy $d_{\mathcal{H}}(P_s, P_t)$ will be discussed.

As the metric distance of distributions $P_s$ and $P_t$, $d_{\mathcal{H}}(P_s, P_t)$ satisfies the following triangle inequality, \emph{i.e.},
\begin{equation}
\begin{split}
d_{\mathcal{H}}(P_s, P_t) \leq d_{\mathcal{H}}(P_s, G(z)) + d_{\mathcal{H}}(P_t, G(z)),
\end{split}
\label{equ:triangle_inequality}
\end{equation}
where $G(z) = \mathcal{N}(0; I)$ is the marginal distribution of $z$.

Recall that two complementary modules $\mathcal{T}_D$ and $\mathcal{T}_F$  (Eq.~\eqref{equ:Mv_Lagrange} and Eq.~\eqref{equ:S_training_adversarial}) alternatively prevent the negative transfer of untransferable knowledge, which encourages the distributions of both $P_s$ and $P_t$ tend to the standard Gaussian, \emph{i.e.}, $P_s \rightarrow\mathcal{N}(0; I)$ and $P_t \rightarrow\mathcal{N}(0; I)$. Consequently, our proposed model forces the last two terms of Eq.~\eqref{equ:triangle_inequality} to be near zero, \emph{i.e.}, $d_{\mathcal{H}}(P_s, G(z))\rightarrow 0$ and $d_{\mathcal{H}}(P_t, G(z))\rightarrow 0$. In summary, our model could efficiently achieve the tighter upper bound for target excepted error $\epsilon^{P_t}(h)$ and reduce domain discrepancy $d_{\mathcal{H}}(P_s, P_t)$.

\begin{figure}[t]
	\begin{minipage}[t]{0.495\linewidth}
		\centering
		\includegraphics[trim = 33mm 73mm 112mm 172mm, clip, height=3.2cm,width=3.8cm]{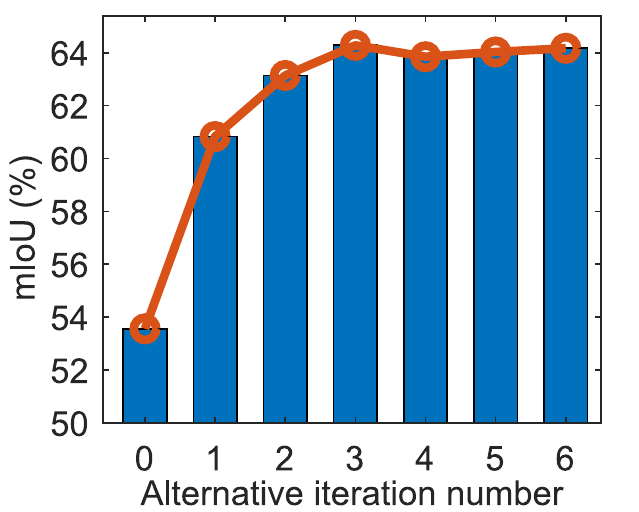}
	\end{minipage}
	\begin{minipage}[t]{0.495\linewidth}
		\centering
		\includegraphics[trim = 33mm 73mm 112mm 172mm, clip, height=3.2cm,width=3.8cm]{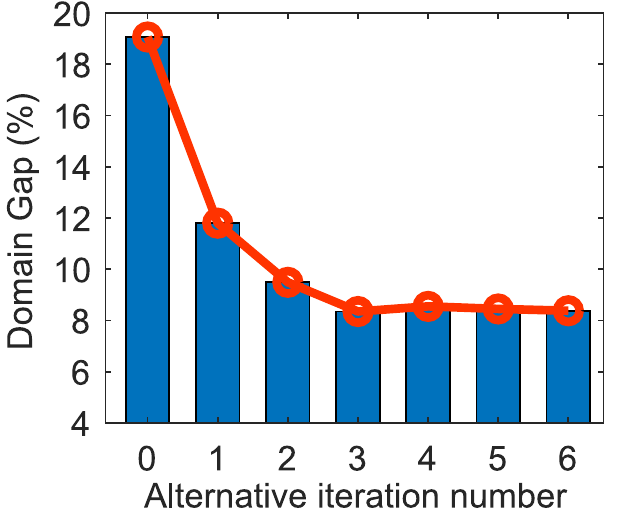}
	\end{minipage}
	\caption{The complementary effect of modules $\mathcal{T}_D$ and $\mathcal{T}_F$ about mIoU (left) and domain gap (right) on the endoscopic dataset.}
	\label{fig:effect_MvMf_medical}
	\vspace{-10pt}
\end{figure}

\section{Experiments}

\subsection{Datasets and Evaluation}	
\textbf{Medical Endoscopic Dataset} \cite{Dong_2019_ICCV} is collected from various endoscopic lesions,  \emph{i.e.}, cancer, polyp, gastritis, ulcer and bleeding. Specifically, it consists of 2969 gasteroscope samples and 690 enteroscopy images. For the training phase, 2969 gastroscope images with pixel annotations are regarded as the source data. We treat 300 enteroscopy samples without pixel labels as target data. In the testing stage, we use the other 390 enteroscopy samples for evaluation. 

\textbf{Cityscapes} \cite{data:city} is a real-world dataset about European urban street scenes, which is collected from 50 cities and has total 34 defined categories. It is composed of three disjoint subsets with 2993, 503 and 1531 images for training, testing and validation, respectively.   	

\textbf{GTA} \cite{data:GTA} consists of 24996 images generated from fictional city scenes of Los Santos in the computer game Grand Theft Auto V. The annotation categories are compatible with the Cityscapes dataset \cite{data:city}.

\textbf{SYNTHIA} \cite{data:synthia} is a large-scale synthetic dataset whose urban scenes are collected from virtual city without corresponding to any realistic city. We utilize its subset called SYNTHIA-RANDCITYSCAPES in our experiments, which contains 9400 images with 12 automatically labeled object classes and some undefined categories.

\textbf{Evaluation Metric:} Intersection over union (IoU) is regarded as basic evaluation metric. Besides, we utilize three derived metrics, \emph{i.e.}, mean IoU (mIoU), IoU of normal ($\mr{IoU}_n$), and IoU of disease ($\mr{IoU}_d$).

\textbf{Notations:} In all experiments, BL represents the baseline network DeepLab-v3 \cite{net:deeplab} without semantic transfer. 

\subsection{Experiments on Medical Endoscopic Dataset}
In our experiments, all the competing methods in Table~\ref{tab:exp_medical_dataset} employ ResNet-101 \cite{net:resnet} as backbone architecture for a fair comparison. From the presented results in Table~\ref{tab:exp_medical_dataset}, we can observe that: 1) Our model could significantly mitigate the domain gap about 11.02\% between source and target datasets when comparing with baseline BL \cite{net:deeplab}. 2) Existing transfer models \cite{Dong_2019_ICCV, Vu_2019_CVPR, Lian_2019_ICCV, Tsai_2019_ICCV, Li_2019_CVPR} perform worse than our model, since they pay equal attention to all semantic representation instead of neglecting irrelevant knowledge, which causes the negative transfer of untransferable knowledge.

\textbf{Effect of Complementary Modules $\mathcal{T}_D$ and $\mathcal{T}_F$:}
This subsection introduces alternative iteration experiments to validate the effectiveness of complementary modules $\mathcal{T}_D$ and $\mathcal{T}_F$. As shown in Figure~\ref{fig:effect_MvMf_medical}, $\mathcal{T}_D$ and $\mathcal{T}_F$ can mutually promote each other and progressively narrow the domain gap along the alternative iteration process. After a few iterations (\emph{e.g.,} the number is 3 for this medical dataset), the performance of our model achieves efficient convergence. After using $\mathcal{T}_D$ to translate transferable visual information, $\mathcal{T}_F$ can further automatically determine how to augment transferable semantic features and in return promote the translation performance of $\mathcal{T}_D$. The experimental results are in accordance with the theoretical analysis in Section~\ref{sec:Theoretical_Analysis}.

\begin{table}[t]
	\centering
	\setlength{\tabcolsep}{1.72mm}
	\caption{Ablation experiments on the medical endoscopic datasets.}
	\scalebox{0.94}{
		\begin{tabular}{|c|cccc|cc|}
			\hline
			Variants & QT & PL & TKB & AA & mIoU(\%) & $\triangle(\%)$ \\
			\hline
			Ours-w/oQT &  & \cmark & \cmark & \cmark & 61.47 & -3.11 \\
			Ours-w/oPL & \cmark &  & \cmark & \cmark & 61.35 & -3.23 \\
			Ours-w/oTKB & \cmark & \cmark &  & \cmark & 62.73 & -1.85 \\
			Ours-w/oAA & \cmark & \cmark & \cmark &  & 63.06 & -1.52 \\
			Ours & \cmark & \cmark & \cmark & \cmark & \textbf{64.58} & - \\
			\hline
		\end{tabular}
	} 				
	\label{tab:exp_ablation_study_medical}
	\vspace{0pt}
\end{table}
\begin{figure}[t]
	\begin{minipage}[t]{0.375\linewidth}
		\centering
		\includegraphics[trim = 20mm 74mm 104mm 155mm, clip, height=2.6cm,width=3.10cm]{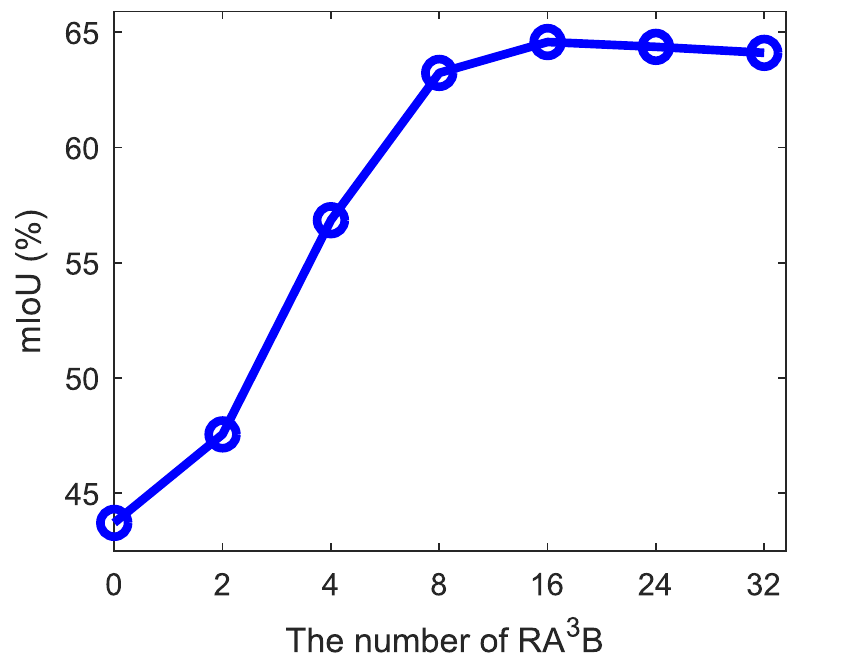}
		\\ {(a) Number of $\mr{RA^2B}$}
		\label{fig:N_RA3B}
	\end{minipage}
	\begin{minipage}[t]{0.65\linewidth}
		\centering
		\includegraphics[trim = 10mm 38mm 15mm 39mm, clip, height=2.6cm,width=5.5cm]{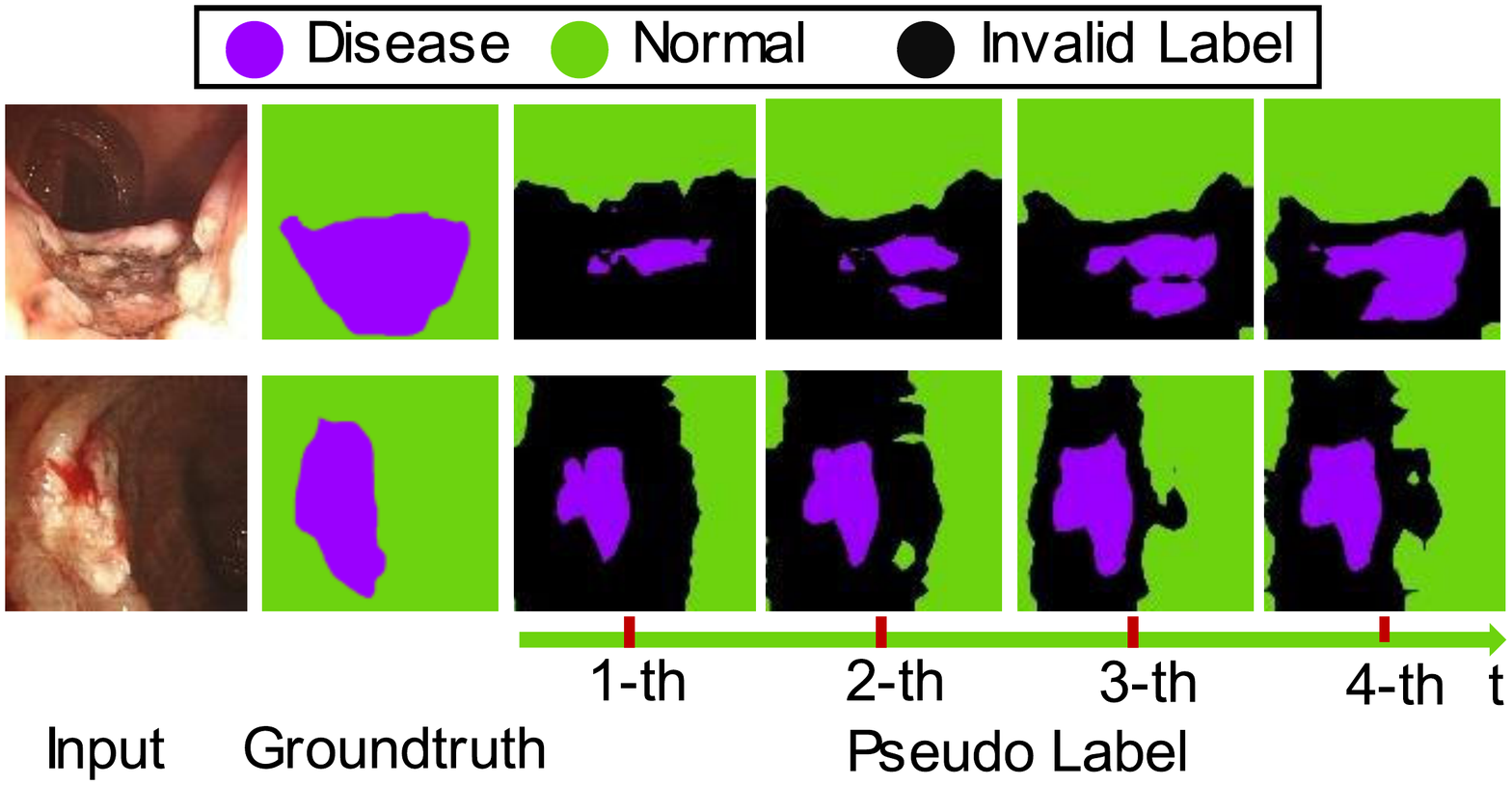}
		\\ {(b) Pseudo labels}
		\label{fig:pseudo_label}
	\end{minipage}
	\vspace{-2pt}
	\caption{The effect of different number of $\mr{RA^2B}$ (left), and the generation process of pseudo labels along the alternative iteration number of modules $\mathcal{T}_D$ and $\mathcal{T}_F$ (right) on the medical dataset.}
	\label{fig:ablation_study}
	\vspace{-5pt}
\end{figure}

\textbf{Ablation Studies:}
To verify the importance of different components in our proposed model, we intend to conduct the variant experiments with the ablation of different components on medical endoscopic dataset, \emph{i.e.}, quantified transferability (QT), pseudo labels (PL), transferability-aware bottleneck (TKB) and attention on attention (AA) of $\mr{RA^2B}$. Training the model without QT, PL, TKB and AA are respectively denoted as Ours-w/oQT, Ours-w/oPL, Ours-w/oTKB and Ours-w/oAA.
From the presented results in Table~\ref{tab:exp_ablation_study_medical}, we can notice that the performance degrades $1.52\% \sim 3.23\%$ after removing any component of our model, which justifies the rationality and effectiveness of each designed component. Besides, with quantified transferability perception from $D_F$, our model could efficiently encode transferable semantic knowledge among source and target datasets while brushing irrelevant representations aside. Multiple $\mr{RA^2B}$s play an essential role in capturing the relevance transferability of transferable knowledge and we set its number as 16, as illustrated in Figure~\ref{fig:ablation_study} (a). Moreover, the distribution shift between different datasets could be further bridged by confident pseudo labels, which are generated progressively along the iteration process, as depicted in Figure~\ref{fig:ablation_study} (b).

\begin{figure}[t]
	\begin{minipage}[t]{0.495\linewidth}
		\centering
		\includegraphics[trim = 20mm 73mm 104mm 155mm, clip, height=3.2cm,width=3.8cm]{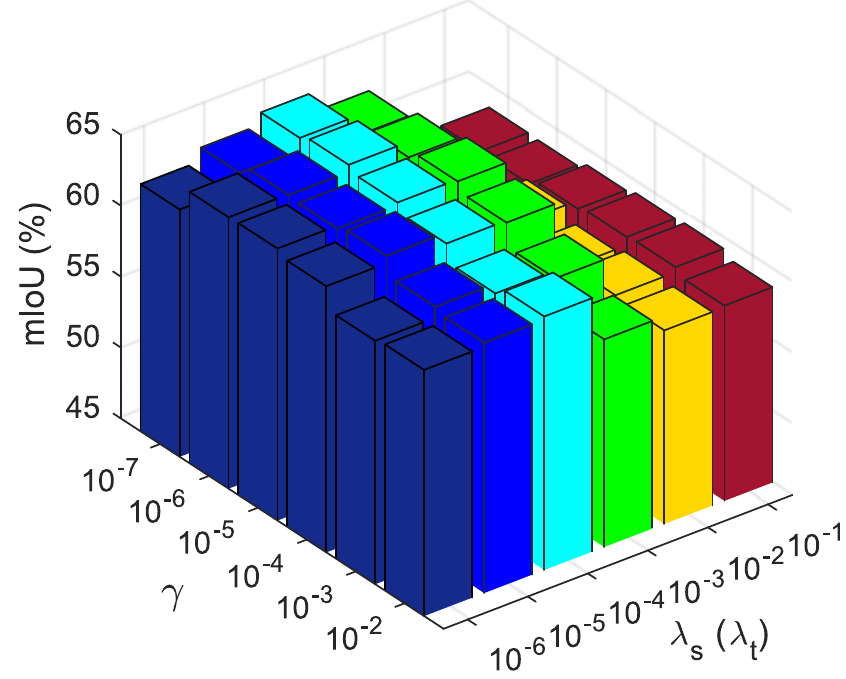}
		\\ {(a) $\alpha=10, T=200$}
		\label{fig:paras_a}
	\end{minipage}
	\begin{minipage}[t]{0.495\linewidth}
		\centering
		\includegraphics[trim = 20mm 73mm 104mm 155mm, clip, height=3.2cm,width=3.8cm]{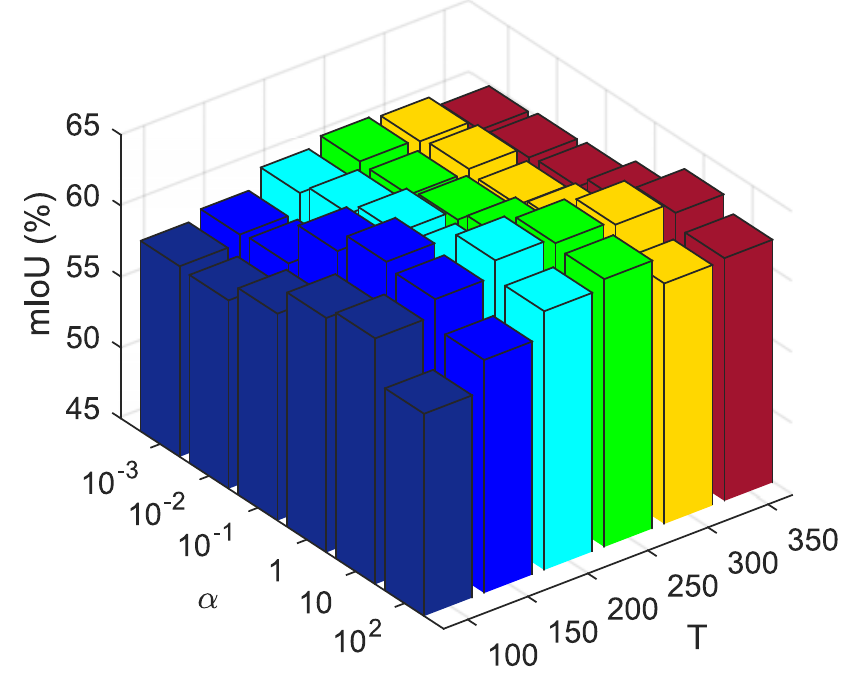}
		\\ {$\!$(b) $\! \gamma = 10^{-6}, \lambda_s(\lambda_t) = 10^{-4}$}
		\label{fig:paras_b}
	\end{minipage}
	\vspace{-12pt}
	\caption{The parameters investigations about $\{\gamma, \lambda_s(\lambda_t)\}$ (left) and $\{\alpha, T\}$ (right) on the medical dataset.}
	\label{fig:exp_params}
	\vspace{-12pt}
\end{figure}

\textbf{Parameters Investigations:}
In this subsection, extensive hyper-parameter experiments are empirically conducted to investigate the effect of hyper-parameters $\{\alpha, T\}$ and $\{\gamma, \lambda_s (\lambda_t)\}$, which assists to determine the optimal parameters. $\lambda_s$ and $\lambda_t$ share same value in our experiments. Notice that our model achieves stable performance over the wide range of different parameters, as shown in Figure~\ref{fig:exp_params}. Furthermore, it also validates that residual transferability-aware bottleneck in Eq.~\eqref{equ:Mv_Lagrange} can efficiently purify the transferable semantic representations with high transfer scores.

\begin{figure}[t]
	\begin{minipage}[t]{0.495\linewidth}
		\centering
		\includegraphics[trim = 34mm 73mm 111mm 167mm, clip, height=3.4cm,width=3.8cm]{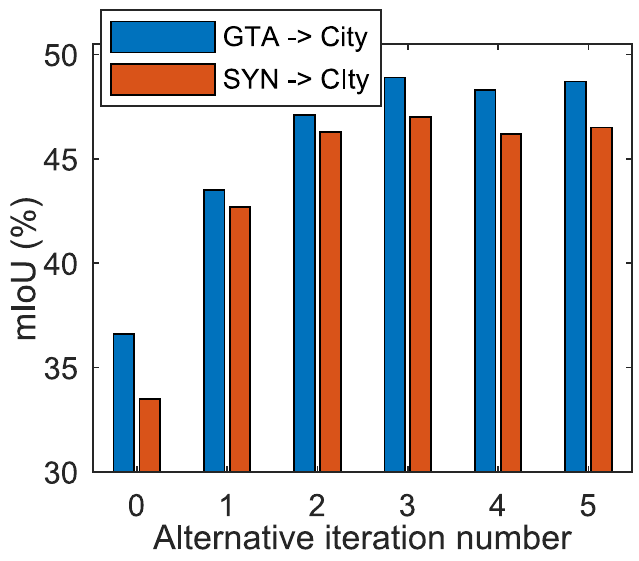}
	\end{minipage}
	\begin{minipage}[t]{0.495\linewidth}
		\centering
		\includegraphics[trim = 34mm 73mm 111mm 167mm, clip, height=3.4cm,width=3.8cm]{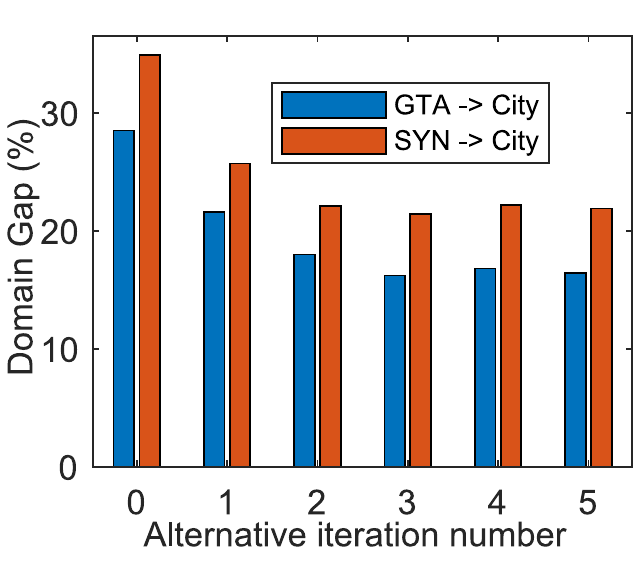}
	\end{minipage}
	\vspace{-15pt}
	\caption{The complementary effect of modules $\mathcal{T}_D$ and $\mathcal{T}_F$ about mIoU (left) and domain gap (right) on several benchmark datasets.}
	\label{fig:effect_MvMf_public} 
	\vspace{-9pt}
\end{figure}

\begin{figure}[t]
	\small
	\centering
	\includegraphics[trim = 16mm 42mm 25mm 57mm, clip, width=240pt, height  =105pt]{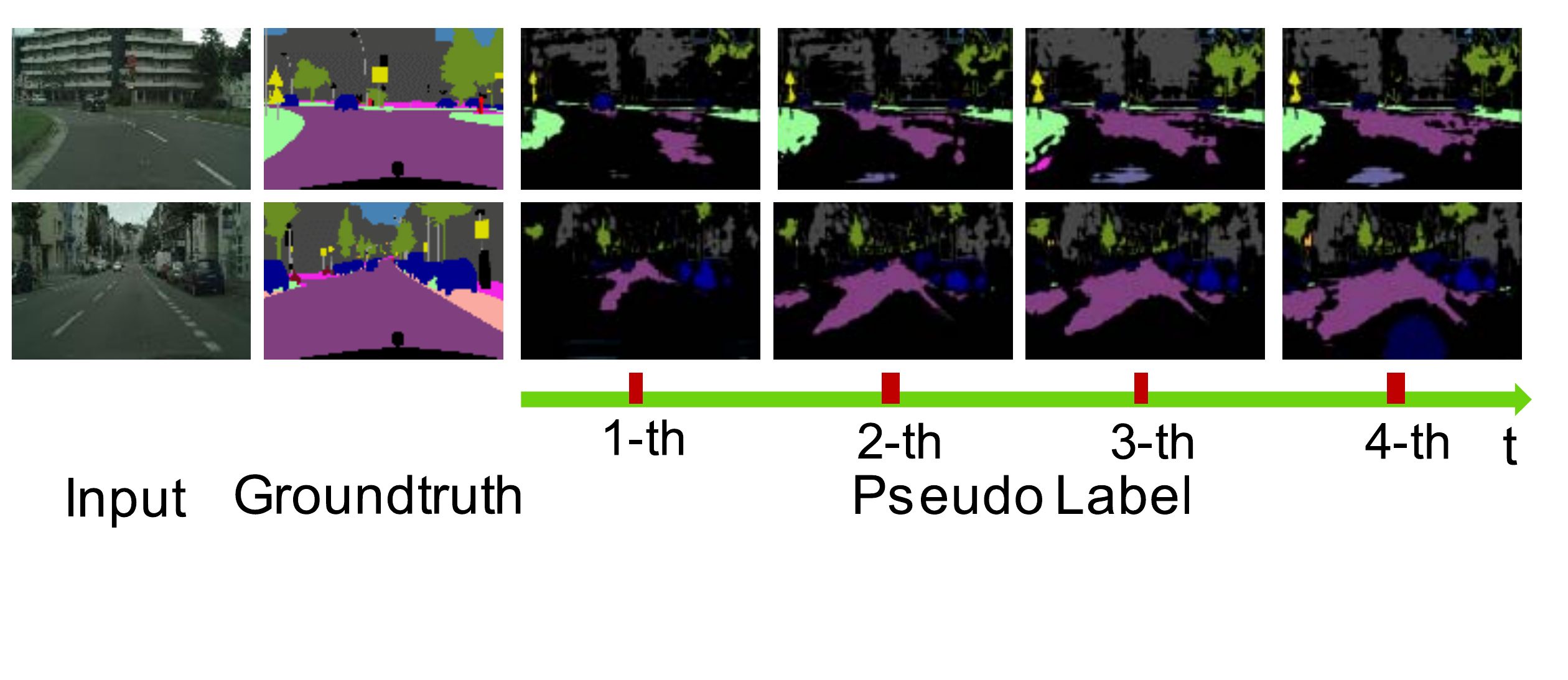} 
	\vspace{-35pt}
	\caption{Pseudo Labels generated along the alternative iteration number of modules $\mathcal{T}_D$ and $\mathcal{T}_F$ on GTA $\rightarrow$ Cityscapes task.}
	\label{fig:pseudo_label_public}
	\vspace{-10pt}
\end{figure}

\begin{table*}[t]
	\centering
	\setlength{\tabcolsep}{1.175mm}
	\caption{Performance comparison of transferring semantic representations from GTA to Cityscapes.}	
	\scalebox{0.772}{
		\begin{tabular}{|c|ccccccccccccccccccc|c|}
			\hline
			Method & road & sidewalk & building & wall & fence & pole & light & sign & veg & terrain & sky & person & rider & car & truck & bus & train & mbike & bike & mIoU(\%)  \\	
			\hline
			\hline
			
%
%
%
			
			LtA \cite{exp:LtA} & 86.5 & 36.0 & 79.9 & 23.4 & 23.3 &23.9 &35.2 &14.8 & 83.4& 33.3 & 75.6 & 58.5 & 27.6 & 73.7 & 32.5 & 35.4 & 3.9 & 30.1 & 28.1 & 42.4 \\
			
			MCD \cite{Saito_2018_CVPR} & 90.3 & 31.0 & 78.5 & 19.7 & 17.3 & 28.6 & 30.9 & 16.1 & 83.7 & 30.0 & 69.1 & 58.5 & 19.6 & 81.5 & 23.8 & 30.0 & 5.7 &25.7 & 14.3 & 39.7  \\
			
			CGAN \cite{exp:CGAN} & 89.2 & 49.0 & 70.7&13.5 & 10.9 & \textcolor{blue}{38.5} & 29.4 & 33.7& 77.9 & 37.6 & 65.8 & \textcolor{red}{\textbf{75.1}} & 32.4& 77.8 & \textcolor{red}{\textbf{39.2}}&45.2 & 0.0& 25.2 & 35.4 & 44.5  \\
			
			
			CBST \cite{Zou_2018_ECCV} & 88.0 & \textcolor{red}{\textbf{56.2}} & 77.0 & 27.4 & 22.4 & \textcolor{red}{\textbf{40.7}} & \textcolor{red}{\textbf{47.3}} & \textcolor{blue}{40.9} & 82.4 & 21.6 & 60.3 & 50.2 & 20.4 & 83.8 & 35.0 & \textcolor{red}{\textbf{51.0}} & 15.2 & 20.6 & 37.0 & 46.2  \\
			
			CLAN \cite{Luo_2019_CVPR} & 87.0 & 27.1 & 79.6 & 27.3 & 23.3 & 28.3 & 35.5 & 24.2 & 83.6 & 27.4 & 74.2 & 58.6 & 28.0 & 76.2 & 33.1 & 36.7 & 6.7 & 31.9 & 31.4 & 43.2  \\
			
			SWD \cite{Lee_2019_CVPR} & 92.0 & 46.4 & 82.4 & 24.8 & 24.0 & 35.1 & 33.4 & 34.2 & 83.6 & 30.4 & 80.9 & 56.9 & 21.9 & 82.0 & 24.4 & 28.7 & 6.1 & 25.0 & 33.6 & 44.5 \\
			
			ADV \cite{Vu_2019_CVPR} & 89.4 & 33.1 & 81.0 & 26.6 & 26.8 & 27.2 & 33.5 & 24.7 & 83.9 & 36.7 & 78.8 & 58.7 & 30.5 & 84.8 & \textcolor{blue}{38.5} & 44.5 & 1.7 & 31.6 & 32.5 & 45.5 \\
			
			BDL \cite{Li_2019_CVPR} & 91.0 & 44.7 & \textcolor{blue}{84.2} & \textcolor{blue}{34.6} & \textcolor{blue}{27.6} & 30.2 &  36.0 & 36.0 & \textcolor{blue}{85.0} & \textcolor{red}{\textbf{43.6}} & 83.0 & 58.6 & 31.6 & 83.3 & 35.3 & 49.7 & 3.3 & 28.8 & 35.6 & \textcolor{blue}{48.5}  \\
			
			SWLS \cite{Dong_2019_ICCV} &\textcolor{red}{\textbf{92.7}} &48.0 &78.8 & 25.7 & 27.2 & 36.0 & \textcolor{blue}{42.2} & \textcolor{red}{\textbf{45.3}} & 80.6 &14.6 &66.0 & 62.1& 30.4 &\textcolor{red}{\textbf{86.2}} & 28.0 &45.6 &\textcolor{red}{\textbf{35.9}} &16.8 & 34.7 & 47.2 \\
			
			DPR \cite{Tsai_2019_ICCV} & \textcolor{blue}{92.3} & \textcolor{blue}{51.9} & 82.1 & 29.2 & 25.1 & 24.5 & 33.8 & 33.0 & 82.4 & 32.8 & 82.2 & 58.6 & 27.2 & 84.3 & 33.4 & 46.3 & 2.2 & 29.5 & 32.3 & 46.5  \\
			
			PyCDA \cite{Lian_2019_ICCV} & 90.5 & 36.3 & \textcolor{red}{\textbf{84.4}} & 32.4 & \textcolor{red}{\textbf{28.7}} & 34.6 & 36.4 & 31.5 & \textcolor{red}{\textbf{86.8}} & 37.9 & 78.5 & \textcolor{blue}{62.3} & 21.5 & \textcolor{blue}{85.6} & 27.9 & 34.8 & \textcolor{blue}{18.0} & 22.9 & \textcolor{red}{\textbf{49.3}} & 47.4 \\
			
			\hline	
			\hline 		
			BL & 75.8 & 16.8 & 77.2 & 12.5 & 21.0 & 25.5 & 30.1 & 20.1 & 81.3 & 24.6 & 70.3 & 53.8 & 26.4 & 49.9 & 17.2 & 25.9 & 6.5 & 25.3 & 36.0 & 36.6  \\  				
			
			Ours-w/oQT & 89.0 & 40.0 & 83.4 & 34.0 & 23.7 & 32.2 & 36.6 & 33.1 & 84.0 & 39.3 & 74.3 & 58.9 & 27.2 & 78.8 & 32.6 & 35.1 & 0.1 & 28.4 & 37.4 & 45.7 \\
			
			Ours-w/oPL & 90.6 & 40.8 & 84.1 & 31.3 & 22.7 & 32.0 & 39.0 & 33.7 & 84.3 & 39.5 & 80.7 & 58.4 & 28.7 & 82.8 & 27.4 & 48.1 & 1.0 & 27.0 & 28.5 & 46.4 \\
			
			Ours-w/oTKB & 88.9 & 45.2 & 82.9 & 32.7 & 26.6 & 31.5 & 34.8 & 34.3 & 83.5 & 38.8 & 81.5 & 60.0 & 31.5 & 80.6 & 30.8 & 44.9 & 5.2 & \textcolor{red}{\textbf{33.8}} & 35.4 & 47.5 \\
			
			Ours-w/oAA & 89.1 & 49.8 & 82.7 & 32.8 & 26.6 & 32.0 & 35.8 & 32.4 & 83.1 & 37.2 & \textcolor{blue}{83.8} & 58.7 & \textcolor{blue}{32.9} & 81.0 & 34.9 &47.1 & 1.5 & 33.1 & \textcolor{blue}{36.8} & 48.0 \\
			
			Ours & 89.4 & 50.1 & 83.9 & \textcolor{red}{\textbf{35.9}} & 27.0 & 32.4 & 38.6 & 37.5 & 84.5 & \textcolor{blue}{39.6} & \textcolor{red}{\textbf{85.7}} & 61.6 & \textbf{\textcolor{red}{33.7}} & 82.2 & 36.0 & \textcolor{blue}{50.4} & 0.3 & \textcolor{blue}{33.6} & 32.1 & \textbf{\textcolor{red}{49.2}}  \\
			
			\hline
			
		\end{tabular}
	}			
	\label{table:exp_gta_cityscapes}
	\vspace{-5pt}
\end{table*}

\begin{table*}[t]	
	\centering
	\setlength{\tabcolsep}{1.33mm}
	\caption{Performance comparison of transferring semantic knowledge from SYNTHIA to Cityscapes.}
	\scalebox{0.865}{
		\begin{tabular}{|c|cccccccccccccccc|c|}
			\hline
			Method & road & sidewalk & building & wall & fence & pole &light &sign & veg & sky & person & rider & car & bus & mbike & bike & mIoU(\%) \\
			\hline			
			\hline				
			
			
			
			LSD \cite{exp:LSD} &80.1 &29.1 &77.5 &2.8 &0.4 &26.8 &11.1 &18.0 &78.1 &76.7 &48.2 & 15.2&70.5 &17.4 &8.7 &16.7 &36.1 \\
			
			
			MCD \cite{Saito_2018_CVPR} & 84.8 & \textcolor{blue}{43.6} & 79.0 & 3.9 & 0.2 & 29.1 & 7.2 & 5.5 & 83.8 & 83.1 & 51.0 & 11.7 & 79.9 & 27.2 & 6.2 & 0.0 & 37.3  \\

			CGAN \cite{exp:CGAN} & \textcolor{blue}{85.0} & 25.8 & 73.5 & 3.4 & \textcolor{red}{\textbf{3.0}} & 31.5 &19.5 & 21.3 & 67.4 & 69.4 & \textcolor{red}{\textbf{68.5}} & 25.0 & 76.5 &  \textcolor{blue}{41.6} & 17.9 & 29.5 & 41.2 \\	
			
			DCAN \cite{Wu_2018_ECCV} & 82.8 & 36.4 & 75.7 & 5.1 & 0.1 & 25.8 & 8.0 & 18.7 & 74.7 & 76.9 & 51.1 & 15.9 & 77.7 & 24.8 & 4.1 & 37.3 & 38.4  \\
			
			CBST \cite{Zou_2018_ECCV} & 53.6 & 23.7 & 75.0 & 12.5 & 0.3 & \textcolor{red}{\textbf{36.4}} & 23.5 & \textcolor{blue}{26.3} & \textcolor{red}{\textbf{84.8}} & 74.7 & \textcolor{blue}{67.2} & 17.5 & \textcolor{blue}{84.5} & 28.4 & 15.2 & \textcolor{red}{\textbf{55.8}} & 42.5  \\
			
			
			
			ADV \cite{Vu_2019_CVPR} & \textcolor{red}{\textbf{85.6}} & 42.2 & 79.7 & 8.7 & 0.4 & 25.9 & 5.4 & 8.1 & 80.4 & 84.1 & 57.9 & 23.8 & 73.3 & 36.4 & 14.2 & 33.0 & 41.2  \\
			
			SWLS \cite{Dong_2019_ICCV} & 68.4 & 30.1 & 74.2 & 21.5 & 0.4 & 29.2 & 29.3 & 25.1 & 80.3 & 81.5 & 63.1 & 16.4 & 75.6 & 13.5 & 26.1 & \textcolor{blue}{51.9} & 42.9  \\
			
			DPR \cite{Tsai_2019_ICCV} & 82.4 & 38.0 & 78.6 & 8.7 & 0.6 & 26.0 & 3.9 & 11.1 & 75.5 & \textcolor{blue}{84.6} & 53.5 & 21.6 & 71.4 & 32.6 & 19.3 & 31.7 & 40.0 \\
			
			PyCDA \cite{Lian_2019_ICCV} & 75.5 & 30.9 & \textcolor{red}{\textbf{83.3}} & 20.8 & 0.7 & \textcolor{blue}{32.7} & 27.3 & \textcolor{red}{\textbf{33.5}} & \textcolor{blue}{84.7} &\textcolor{red}{\textbf{85.0}} & 64.1 & \textcolor{blue}{25.4} & \textcolor{red}{\textbf{85.0}} & \textcolor{red}{\textbf{45.2}} & 21.2 & 32.0 & \textcolor{blue}{46.7}  \\
			
			\hline
			\hline		
			BL & 55.6 & 23.8 & 74.6 & 9.2 & 0.2 & 24.4 & 6.1 & 12.1 & 74.8 & 79.0 & 55.3 & 19.1 & 39.6 & 23.3 & 13.7 & 25.0 & 33.5  \\				
			
			Ours-w/oQT & 69.4 & 30.9 & 79.8 & 21.3 & 0.5 & 30.2 & \textcolor{blue}{31.0} & 22.7 & 82.3 & 82.6 & 66.4 & 15.2 & 79.1 & 20.5 & \textcolor{red}{\textbf{26.7}} & 48.2 & 44.2  \\
			
			Ours-w/oPL & 70.3 & 32.1 & 77.8 & \textcolor{red}{\textbf{22.9}} & \textcolor{blue}{0.8} & 29.6 & \textcolor{red}{\textbf{32.4}} & 24.3 & 81.7 & 80.1 & 62.9 & 22.0 & 75.4 & 26.2 & 25.3 & 51.0 & 44.7  \\
			
			Ours-w/oTKB & 78.6 & 39.2 & \textcolor{blue}{80.4} & 19.5 & 0.6 & 27.8 & 29.1 & 21.5 & 80.8 & 82.0 & 64.5 & 24.7 & 83.5 & 29.6 & 24.1 & 46.3 & 45.8 \\
			
			Ours-w/oAA & 81.3 & 41.5 & 79.2 & 21.8 & 0.7 & 28.3 & 27.6 & 20.1 & 81.7 & 80.9 & 62.7 & 25.3 & 82.1 & 34.5 & 23.6 & 47.3 & 46.2 \\
			
			Ours & 81.7 & \textcolor{red}{\textbf{43.8}} & 80.1 & \textcolor{blue}{22.3} & 0.5 & 29.4 & 28.6 & 21.2 & 83.4 & 82.3 & 63.1 & \textcolor{red}{\textbf{26.2}} & 83.7 & 34.9 &  \textcolor{blue}{26.3} & 48.4 & \textcolor{red}{\textbf{47.2}} \\ 
			\hline 		
		\end{tabular}
	}		
	\label{table:exp_synthia_cityscapes}
	\vspace{-7pt}
\end{table*}

\subsection{Experiments on Benchmark Datasets}
Extensive experiments on several non-medical benchmark datasets are also conducted to further illustrate the generalization performance of our model. For a fair comparison, we set the same experimental data configuration with all comparable state-of-the-arts \cite{exp:CGAN, exp:LtA, Dong_2019_ICCV, Lee_2019_CVPR, Vu_2019_CVPR}. To be specific, in the training phase, GTA \cite{data:GTA} and SYNTHIA \cite{data:synthia} are regarded as the source dataset, and the training subset of Cityscapes \cite{data:city} is treated as the target dataset. We use the validation subset of Cityscapes \cite{data:city} for evaluation.  Table~\ref{table:exp_gta_cityscapes} and  Table~\ref{table:exp_synthia_cityscapes} respectively report the results of transferring from GTA and SYNTHIA to Cityscapes.
From Table~\ref{table:exp_gta_cityscapes} and  Table~\ref{table:exp_synthia_cityscapes}, we have the following observations: 1) Our model outperforms all the existing advanced transfer models \cite{exp:CGAN, exp:LtA, Dong_2019_ICCV, Saito_2018_CVPR, Vu_2019_CVPR} about $0.5\% \sim 11.1\%$, since two complementary modules could alternatively explore where and how to highlight transferable knowledge to bridge the domain gap, as shown in Figure~\ref{fig:effect_MvMf_public}. 2) Ablation studies about different components illustrate they play an important role in highlighting transferable domain-invariant knowledge to improve the transfer performance. 3) Our model achieves larger improvements for those hard-to-transfer classes with various appearances among different datasets (\emph{e.g.}, sidewalk, wall, motorbike, rider, sky and terrain) by selectively neglecting untransferable knowledge. In addition, Figure~\ref{fig:pseudo_label_public} presents the iteratively generated pseudo labels on GTA $\rightarrow$ Cityscapes task, which narrows the distribution divergence.

\section{Conclusion}
In this paper, we develop a new unsupervised semantic transfer model including two complementary modules ($\mathcal{T}_D$ and $\mathcal{T}_F$), which alternatively explores transferable domain-invariant knowledge between labeled source gastroscope lesions dataset and unlabeled target enteroscopy diseases dataset. Specifically, $\mathcal{T}_D$ explores where to translate transferable visual characterizations while preventing untransferable translation. $\mathcal{T}_F$ highlights how to augment those semantic representations with high transferability scores, which in return promotes the translation performance of $\mathcal{T}_D$. Comprehensive theory analysis and experiments on the medical endoscopic dataset and several non-medical benchmark datasets validate the effectiveness of our model.

{\small
\bibliographystyle{ieee_fullname}
\bibliography{UnsupervisedLesionsTransfer}
}

\end{document}